\documentclass[final]{cvpr}

\makeatletter
\@namedef{ver@everyshi.sty}{}
\makeatother

\usepackage[pagebackref=true,breaklinks=true,colorlinks,bookmarks=false]{hyperref}

\usepackage{pgfplots}

\usepackage{times}
\usepackage{epsfig}
\usepackage{graphicx}
\usepackage{amsmath}
\usepackage{amssymb}
\usetikzlibrary{calc, positioning}
\usepackage{array}
\usepackage{eqparbox}

\usepackage{float}
\usepackage{xifthen,bm,tikz,tikz-cd,adjustbox,afterpage,algorithmic}
\usetikzlibrary{patterns}

\usepackage{caption}
\usepackage{subcaption}
\usepackage{multirow}
\usepackage{booktabs}
\usepackage{arydshln}

\usepackage[ruled,vlined]{algorithm2e}
\usepackage{amsfonts}
\SetAlFnt{\footnotesize}
\SetKwInput{KwInit}{Initialization}

\definecolor{green1}{HTML}{004c00}
\definecolor{green2}{HTML}{009900}
\definecolor{green3}{HTML}{47da47}
\definecolor{green4}{HTML}{99ea99}
\definecolor{blue1}{HTML}{003566}
\definecolor{blue2}{HTML}{186fc0}
\definecolor{blue3}{HTML}{68aae7}
\definecolor{blue4}{HTML}{a1caf0}
\definecolor{red0}{HTML}{E74C4C}
\definecolor{red1}{HTML}{990000}
\definecolor{red2}{HTML}{DD0000}
\definecolor{red3}{HTML}{E74C4C}
\definecolor{red4}{HTML}{F19999}
\definecolor{yellow1}{HTML}{F1C239}
\definecolor{gray1}{HTML}{C8C8C8}
\definecolor{gray2}{HTML}{707070}
\definecolor{gray3}{HTML}{505050}
\definecolor{brown1}{HTML}{654321}

\def\cvprPaperID{56} % *** Enter the CVPR Paper ID here
\def\confYear{CVPR 2021}

\begin{document}

%%%%%%%%% TITLE
\title{Class-Incremental Experience Replay for\\Continual Learning under Concept Drift}

\author{\L{}ukasz Korycki\\
Department of Computer Science\\
Virginia Commonwealth University\\
Richmond, VA, USA\\
{\tt\small koryckil@vcu.edu}
\and
Bartosz Krawczyk\\
Department of Computer Science\\
Virginia Commonwealth University\\
Richmond, VA, USA\\
{\tt\small bkrawczyk@vcu.edu}
}

\maketitle

%%%%%%%%% ABSTRACT
\begin{abstract}
Modern machine learning systems need to be able to cope with constantly arriving and changing data. Two main areas of research dealing with such scenarios are continual learning and data stream mining. Continual learning focuses on accumulating knowledge and avoiding forgetting, assuming information once learned should be stored. Data stream mining focuses on adaptation to concept drift and discarding outdated information, assuming that only the most recent data is relevant. While these two areas are mainly being developed in separation, they offer complementary views on the problem of learning from dynamic data. There is a need for unifying them, by offering architectures capable of both learning and storing new information, as well as revisiting and adapting to changes in previously seen concepts. We propose a novel continual learning approach that can handle both tasks. Our experience replay method is fueled by a centroid-driven memory storing diverse instances of incrementally arriving classes. This is enhanced with a reactive subspace buffer that tracks concept drift occurrences in previously seen classes and adapts clusters accordingly. The proposed architecture is thus capable of both remembering valid and forgetting outdated information, offering a holistic framework for continual learning under concept drift. 
\end{abstract}

%%%%%%%%% BODY TEXT
\section{Introduction}
\label{sec:int}

Contemporary real-world problems generate challenging and ever-growing data with dynamic properties. This kickstarted exciting developments of novel machine learning algorithms capable of constant accumulation of new information \cite{Aljundi:2019}, aggregating useful data \cite{Sahoo:2018}, and handling its non-stationary properties \cite{Lu:2019}. Two fields are being developed in parallel -- continual learning \cite{Parisi:2019} and data stream mining \cite{Krawczyk:2017}. The former focuses on how to retain useful knowledge within the model, while allowing its growth and accumulation of new information. The latter focuses on adaptation to the current state of data, detecting the phenomenon known as concept drift, and swift adaptation to any changes taking place \cite{Lu:2019}. One must notice that those two approaches offer complementary views on the problem of continual learning from dynamic data and thus should be bridged together, leading us to develop robust and adaptive learning algorithms. 

\smallskip
\noindent \textbf{Research hypothesis.} Class-incremental continual learning can be effectively extended to allow at the same time: (i) avoiding catastrophic forgetting by effectively accumulating knowledge from new classes; and (ii) monitoring changes in previously learned classes (revisiting) with automatic adaptation to concept drift.

\begin{figure*}[h]
\leftskip10pt\includegraphics[width=.95\textwidth]{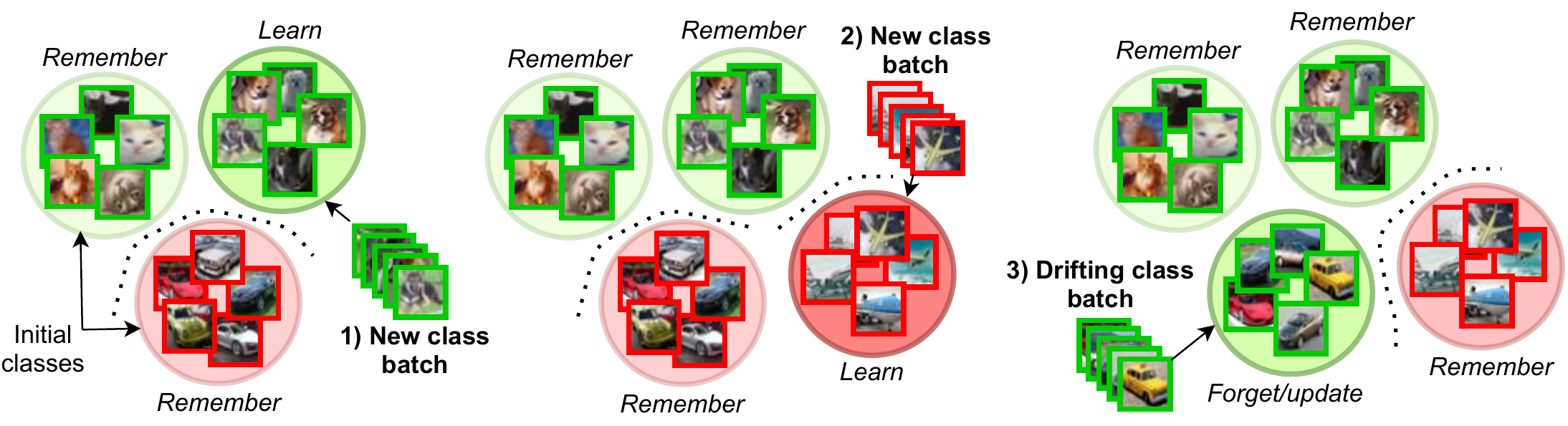}
\caption{Three vital aspects of a holistic approach to continual learning: learning new classes, retaining previous knowledge, and adapting to concept drifts, illustrated by the example of a binary recommendation system (\textcolor{green2}{like} or \textcolor{red2}{dislike}).}
\label{fig:fram}
\end{figure*}

\smallskip
\noindent \textbf{Motivation.} Existing continual learning methods assume that the once learned knowledge should be remembered and stored in the model. This assumes that information once learned stays permanently valid. This is not true, as modern dynamic data sources may be affected by concept drift, thus changing the properties of some classes, as depicted by the example of a binary recommendation system given in Fig.~\ref{fig:fram}. This calls for developing methods that can bridge the gap between continual learning (knowledge retaining) and data stream mining (concept drift adaptation). 

\smallskip
\noindent \textbf{Overview.} We propose a holistic approach to class-incremental continual learning, based on experience replay. The novelty of our work is that our algorithm allows for both avoiding catastrophic forgetting and automatic update of previously learned classes if they are affected by concept drift. We distinguish three vital aspects of the proposed framework: (i) capability for class-instrumental continual learning; (ii) capability for retaining useful knowledge to mitigate catastrophic forgetting; and (iii) capability of adaptation to changes by forgetting outdated knowledge and updating the model. Our approach combines centroid-driven memory for storing class-based prototypes with a reactive subspace buffer that can detect and react to concept drift affecting one or more classes. It traces the dominant class in each of the clusters, allowing for switching labels among clusters and splitting them whenever local changes are being detected. We simultaneously ensure the diversity of information stored within class buffers with their reactivity to concept drifts. 

\smallskip
\noindent \textbf{Main contributions.} Our work offers the following advancements to the field of continual learning.

\begin{itemize}
	\item \textbf{Combination of aggregation and adaptation.} We propose a unifying view on continual learning that combines avoiding catastrophic forgetting during class-incremental learning with adaptation to changes affecting previously learned classes.
	\item \textbf{Reactive subspace buffer.} We develop a novel experience replay approach that combines clustering-driven buffers for managing data diversity with cluster tracking, switching, and splitting for forgetting outdated information and automatic adaption to concept drift without a need for explicit change point detection. 
	\item \textbf{Realistic continual learning scenario under concept drift.} We discuss a realistic and illustrative learning scenario -- continual preference learning and recommendation. As both users may acquire new preferences and their old preferences may change over time, this problem touches both avoiding catastrophic forgetting in incremental learning and the need for handling concept drift.
\end{itemize}

\section{Related works}
\label{sec:rel}

\noindent \textbf{Continual learning -- robustness to forgetting.} The field of continual learning focuses on incremental incorporation of new information into the model, while preserving the knowledge learned on previous classes or tasks \cite{Parisi:2019}. This leads to two challenges: how to grow model to make space for new knowledge and how to avoid catastrophic forgetting. Model growing is mainly discussed from the perspective of deep neural networks, where recent works suggest to either elastically add new neurons to the network \cite{Li:2019}, have a pretrained bigger structure that is incrementally populated \cite{Sokar:2020}, or use hypernetworks to control the accumulation of new data \cite{Oswald:2020}. 

At the same time, while the model is accumulating new classes/tasks it may become biased towards this recent distribution \cite{Wu:2019}. Continual learning models must be robust to catastrophic forgetting, to maintain high accuracy in previous classes. Most common solutions to this problem use instance buffers from previous classes for experience replay \cite{Fujimoto:2020}, specialized parts of neural networks for solving each task while freezing the rest with masking \cite{Mallya:2018}, or use regularization and parametrization methods to debias the learning procedure \cite{Chen:2020, Kirkpatrick:2017}. 

\smallskip
\noindent \textbf{Data stream mining -- robustness to changes.} The field of data stream mining focuses on continuous adaptation to the newest incoming data, with the assumption that they are the most representative \cite{Krawczyk:2017}. This is dictated by the non-stationary data characteristics and presence of concept drift that may affect class boundaries, distributions, and features \cite{Korycki:2020, Lu:2019}. Concept drift can be handled in either an explicit or implicit manner. The former approach uses drift detectors -- external monitoring tools that evaluate selected properties of incoming instances and/or learning models to signal the moment of a drift \cite{Barros:2018}. Once the drift is detected, the learning model is replaced with a new one trained on most recent data, thus facilitating forgetting of outdated concepts. The letter approach uses adaptive learning techniques, assuming that the underlying model will smoothly follow the changes in a data stream. Here sliding windows \cite{Gou:2020} and online classifiers \cite{Cano:2020} are the most popular solutions, allowing for learning from incoming instances, while simultaneously forgetting the old information.

\section{Unifying continual learning with concept drift adaptation}
\label{sec:uni}

\noindent \textbf{Need for unifying approach to continual learning.} Both discussed approaches to learning from dynamic data focus on different aspects and are being developed in relative separation. One should notice that both raise vital issues that are present in many dynamic real-world problems \cite{Yu:2020,Zhang:2019sd}. This allows us to postulate the need for a unifying approach that will bridge those two domains and offer a holistic view on the continual learning paradigm \cite{Caccia:2020}. The questions that all continual learning models should answer are (i) how to accumulate new information and expand the capabilities of a learning model; (ii) how to remember important and useful information over time; and (iii) how to detect and forget outdated concepts. This proposed unified approach will allow continual learning models to offer leveraged robustness to both catastrophic forgetting and concept drift, making them a step closer to solving emerging real-world problems.

\smallskip
\noindent \textbf{Realistic scenario for continual learning under concept drift.} Most works on concept drift use artificial data generators \cite{Sahoo:2018} or forcefully inject drift into every class without considering the reason behind it (e.g., \textit{apples} changing into \textit{cars}). Those examples are not very convincing and thus there is a need for finding a way of creating concept drift problems that are rooted in reality. We propose to focus on user's preferences, as they are an excellent example of continual and streaming data, as depicted in Fig.~\ref{fig:fram}. 

Users are constantly processing new information given to us from social media, the internet, or news outlets, learning about new things we have not seen before. Those new things may become interesting to a user or not – but they still need to be processed in a continuous manner, calling for class-incremental mechanisms. A new topic does not become the major or only interest for the user; thus it cannot overshadow the previously seen ones. Therefore, catastrophic forgetting must be avoided to retain not only the most current, but all topics relevant to a given user.  At the same time, our preferences and tastes are not static. We change our interests within the span of years, months, or even days. A concept that was interesting to the user at a given point cannot be assumed as interesting indefinitely. A continual learning system must thus be able of revisiting previously learned knowledge and updating it according to any shifts in preferences. This calls for concept drift adaptation approaches, as previously seen topics may evolve over time and the interest of users in them may either increase or decrease over time. Creating a true continual learning system over user preferences is a real-world and practical illustration explaining the need for a holistic approach capable of remembering new concepts and selective forgetting with adaptation to changes in the old ones.

\section{Class-incremental experience replay under concept drift}
\label{sec:erc}

The prevalent majority of the class-incremental methods based on experience replay focus on storing the most representative instances or prototypes for stationary data \cite{Rolnick:2019}. They rely on the assumption that classes of the observed and selected instances cannot change, therefore there is no need to control them. As a result, the instances picked for a given class will remain in its buffer for a very long time and the only criterion which may trigger their removal or replacement will be representativeness or diversity of the memory \cite{Foerster:2017}. However, in many real-world applications the mentioned assumption does not hold true. In the presented example of a binary recommendation system, the preferences may change, invalidating some of the experiences stored in the buffer. In such a case, we have to address the concept drift problem and update our memory adequately. 

In the following sections, we introduce two commonly used basic algorithms -- class buffers and centroid-driven memory -- in the context of the given problem. We also propose an adaptive experience replay approach capable of adapting to concept changes.

\subsection{Class buffers}
Standard experience replay methods tackle the catastrophic forgetting problem by storing a separate buffer per class. They assume that a label of an incoming instance is known, so they can perfectly balance the storage and reasonably diversify their available memory. Due to practical concerns, the class buffers have limited capacity, therefore, there is the necessity of selecting which (or if) previously captured instances should be replaced with the currently incoming ones. 

The most simplistic approaches use basic algorithms like FIFO (queue) effectively acting as sliding windows \cite{Schaul:2016}. The problem with such methods is that they may very quickly erase the memory of earlier examples, that may be representative for a given problem, leading to catastrophic forgetting \cite{Rostami:2019}. Assuming that incoming instances may be somehow correlated in time, one possible modification mitigating the issue is to enforce a wider spread of the stored instances across time. To achieve that, we can sample instances stored in the buffer, using a simple formula:
\begin{equation}
r \sim \mathcal{U}(0,1) < \tau,
\end{equation}

\noindent where $r$ is a random variable sampled from the uniform distribution and $\tau$ is a threshold specified by a user. By increasing the threshold we can enforce quicker replacements, while, on the other hand, by decreasing it we can make the buffer more conservative. Too low $\tau$ will lead to impaired learning from new data, while too high $\tau$ will inevitably lead to catastrophic forgetting. Although balanced thresholds should be preferable for classic stationary scenarios, such approaches will fail when concept drift occurs, imposing unnecessary avoidance of forgetting and impeding adaptation to changes.

\subsection{Centroid-driven memory}
Usually, the simple class buffer methods are too simplistic, since they do not utilize any significant characteristics of the incoming data. This is especially important, when we have to deal with complex or high-level abstract classes (e.g. in recommendation), since it is very likely there are several different subspaces that should be covered by the maintained buffers. This is where the clustering methods can be found very useful \cite{Sublime:2017}. They are usually utilized to diversify the replay buffer by grouping instances into differing groups, which should result in a better coverage of the decision space \cite{Lee:2016}. The centroids can be used as instances themselves (prototypes) \cite{Hayes:2019, Shi:2018}, or solely as representations of buffers to forward new examples to their similar memory cells \cite{Zhang:2019}. In this work, we focus on the latter approach.

Although the centroid-driven approaches are one step further than the simple class buffers, they are still susceptible to concept drifts. The reason for that is the fact that they very often do not check whether previously created clusters are still valid for a given class. If a given subconcept cluster changes its label (e.g., from liking to disliking), the new incoming instances will start (slowly) updating class centroids for another class. However, they will not affect the old cluster for the previous class, since the new instances will not be identified as those belonging to it, leaving it obsolete and impeding the learning process when one samples from it. This will, once again, lead to the opposite of catastrophic forgetting, resulting in much slower or non-existent adaptation to the current concepts.

\subsection{Reactive subspace buffer}

To address the presented problem, we propose a modification of the clustering-driven replay buffers, called \textbf{Reactive Subspace Buffer (RSB)}, capable not only of efficient knowledge aggregation, but also of adequate forgetting when it is needed. The outline of the algorithm is given in Algorithm~\ref{alg:framework}. More details can be found in our public repository: \href{https://github.com/lkorycki/rsb}{github.com/lkorycki/rsb}.

\begin{algorithm}[h]
	\BlankLine
	\KwData{min centroids $c_{max}$, max centroids $c_{max}$, buffer size $b_{max}$, window size $\omega_{max}$}
	\KwResult{replay buffers $B$ at every iteration}
	\Repeat{stream ends}{
		receive incoming instance $x$ and its label $y$\;
		\uIf{$c_y < c_{min}$}{
			add new centroid $C_{new}$, buffer $B_{new}$ and window $W_{new}$ for $y$\;
			continue\;
		}
		\BlankLine
		find the closest centroid $C_{x}$ for $x$\;
		\BlankLine
		\uIf{$y_{C_{x}} == y$}{
			update centroid $C_x$, buffer $B_x$ and its window $W_x$ with $(x, y)$\;
		}\uElseIf{$x$ is within $C_x$}{
			update window $W_x$ with $(x, y)$\;
			\BlankLine
			\uIf{should switch $C_x$}{
				move $C_x$ to centroids of $y$ and update it using $W_x$\;
			}
		}\uElse{
			find the closest centroid $C_{y,x}$ for $(x, y)$\;
			\BlankLine
			\uIf{$x$ is within $C_{y,x}$ {\normalfont or} $c_y \geq c_{max}$}{
				update centroid $C_{y,x}$, buffer $B_{y,x}$ and its window $W_{y,x}$ with $(x, y)$\;
			}\uElse{
				add new centroid $C_{new}$, buffer $B_{new}$ and window $W_{new}$ for $y$\;
			}
		}
		
		\BlankLine
		check for splits and removals
	}
	\caption{Reactive Subspace Buffer (RSB).}
	\label{alg:framework}
\end{algorithm}

In the given algorithm, for each new instance $x$ with a label $y$, we first ensure that there are at least $c_{min}$ centroids for the class. Then, we find the nearest cluster $C_x$ for the given instance $x$. If the given cluster belongs to the class $y$ of the instance, we simply update it, its buffer $B_x$ of maximum size $b_{max}$ and sliding window $W_x$ of maximum size $\omega_{max}$, where the last component is responsible for tracking the most current concepts for the given centroid. Otherwise, there is a risk that a concept drift appeared and instances of a different class have started appearing around the centroid. Therefore, if the instance $x$ is sufficiently close (we use simple standard deviation rules), we update the sliding window of the centroid $C_x$, but not the cluster itself. Now, if we detect that there is a significant number of instances with labels different from the current label of the centroid, we switch it to the new majority class. By doing so, we allow the buffer to quickly react to a potential drift. Otherwise, we find the closest centroid $C_{y, x}$ belonging to the same class $y$ as $x$ and we either update it, if $x$ is sufficiently close to the cluster and the maximum number of clusters $c_{max}$ has not been reached, or create a new centroid for the given class $y$.

Finally, for each centroid $C$, after every $n_s$-th update of its sliding window $W$, we check whether it did not switch labels but is impure enough to be split into two separate classes. We apply a simple formula checking if $c_1 / c_2 - 1.0 < \tau_s$, where $c_1$ and $c_2$ are the first and second most numerous classes in the cluster and $\tau_s$ is a threshold determined by a user. During this step we also get rid of minuscule clusters for which less than $\tau_r=\alpha_r\omega_{max}$ instances were registered, where $\alpha_r$ is set by a user.

The whole algorithm is then used as a part of the experience replay method, in which we attempt to sample one instance for each centroid $C$ from its buffer $B$, based on the purity criterion:
\begin{equation}
\gamma_C = \tanh(\beta \frac{c_1 - c_2}{c_1 + c_2}) > r \sim \mathcal{U}(0,1),
\end{equation}

\noindent where $c_1$ and $c_2$ are, once again, the most numerous classes in the cluster, and $\beta=4$. By doing so, we provide an additional mechanism preventing us from enhancing outdated or at least uncertain concepts. Finally, since by using probabilistic sampling we make the total number of sampled instances non-deterministic, we apply oversampling to balance the selected batch.

To summarize -- by enabling: (i) tracking the current dominant classes in a given cluster, (ii) switching labels between clusters, (iii) splitting them, and (iv) sampling from the replay buffer based on clusters purity, we make the centroid-driven algorithm sensitive to concept changes. At the same time, by maintaining stable replay buffers for subspaces that do not change, we can still avoid catastrophic forgetting. As a result, we are able to obtain a method capable of both remembering what is valid and forgetting what is outdated. In addition, since our method is based on local buffers, it should be able to efficiently diversify more complex concepts without explicit knowledge of its subconcepts.

\section{Experimental study}
\label{sec:exp}

In the experimental study, we attempt to prove that our algorithm is capable of class-incremental learning from stationary and non-stationary data. We aim to show that it can both (i) avoid catastrophic forgetting by maintaining diversified subspace-oriented replay buffers, and (ii) adapt to concept drifts by forgetting outdated information. All of the presented experiments can be conducted using scripts provided in the mentioned repository.

\subsection{Data}
To evaluate the proposed algorithm we decided to simulate a binary recommendation system by assigning superclasses (0/1) to the classes from original data sets. By doing so we could simulate the situation in which a user likes or dislikes certain types of available images (subconcepts). We constructed two types of class-incremental data sets: \textbf{stationary} and \textbf{drifting}. 

\begin{figure}[h]
\includegraphics[width=\columnwidth]{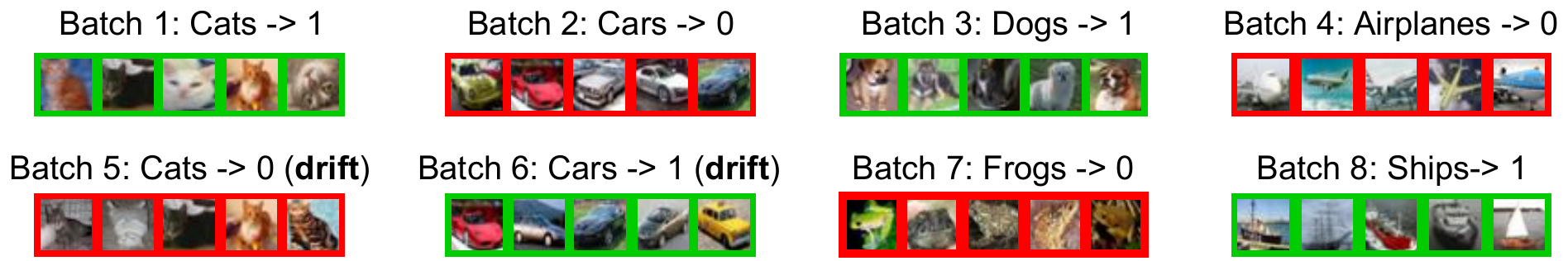}
\caption{The general idea of the design for the drifting benchmark sequences.}
\label{fig:des}
\end{figure}

For the former, we simply used five image benchmarks: MNIST, FASHION, SVHN, CIFAR10 and IMAGENET10, which is a subset of the 64x64 ImageNet set. During the evaluation we were feeding our models class after class, interleaving 0/1 assignments (for example, the first class from CIFAR10 was 1, the second one was 0, and so on). For the latter scenario, we were changing the 0/1 labels for two consecutive classes after three or four stationary ones. As a result, we obtained 30 batches of classes for each data set, representing both stationary and concept drifting periods (for more details please refer to the repository). An example of our approach is depicted in Fig.~\ref{fig:des}.

\subsection{Algorithms}
We evaluated our algorithm as a part of the experience replay framework. The module consisted of a classifier (a neural net) and the replay buffer, which was used to sample additional instances for a given input batch. In order to compare our method (\textbf{RSB}) with other mentioned approaches, we run experiments using four additional classifiers: (i) offline neural network retraining after each batch (\textbf{OFFLINE}), (ii) a naively fine-tuning neural network (\textbf{NN}), which learned from the batches without any additional mechanisms for handling catastrophic forgetting, (iii) a simple class buffer (\textbf{CB}), which stored separate buffers for both recommendation classes, and a centroid-based method that utilized an on-line k-means algorithm to create representations of the original classes treated as subconcepts (or subspaces) of the recommendation space (\textbf{SB}).

While configuring our method, we used the following values of its parameters: $c_{min} = 0.5 c_{max}$, where $c_{max}=10$ for all of the data sets except for FASHION for which we set $c_{max}=20$ based on preliminary experiments. Each buffer of the method could store at most $b_{max}=100$ instances and equal was the size of each sliding window $\omega_{max}=100$. We also empirically set $n_s=1000$ and $\tau_s=0.5$ for splitting, and $\alpha_r=0.4$ for removals. These settings worked very well with all of the considered data sets. We used the same values of $c_{max}$ and $b_{max}$ for the SB algorithm. When it comes to the CB method, we set $b_{max} = 2000$ per class to provide similar memory resources compared with RSB and SB. Furthermore, we distinguished CB with $\tau = 0.0$ (\textbf{CB0}) and $\tau = 1.0$ (\textbf{CB1}) to check their performance in stationary and non-stationary scenarios.

All of the mentioned algorithms used pretrained convolutional feature extractors, from which we used representations returned by a middle layer of the classifier (we needed a high-level representation due to the nature of our task). For MNIST and FASHION we used a simple CNN with two convolutional layers consisting of 32 (5x5) and 64 (3x3) filters, interleaved with ReLU, batch normalization and max poolig (2x2). For SVHN, CIFAR10 and IMAGENET10 we utilized ResNet18. As a trainable classifier we chose a 3-layer fully connected net with 512, 256, 128 neurons in the hidden layers interleaved with ReLU, batch normalization and dropout ($p=0.5$). During training we used the Adam optimizer. After each batch, the classifier learned for either 5 epochs (IMAGENET10) or 10 (the rest). Additionally, we initialized each algorithm with 10\% of the first and the second class.

\subsection{Evaluation}
We evaluated the presented methods in a class-incremental setting, where each original class is treated as a subconcept of the binary recommendation space and comes as a whole in a form of a batch (Fig. \ref{fig:des}). In our scenario, we assume that old subconcepts may become outdated and batches may change their labels. We measured the accuracy of a given algorithm after each batch (a new or updated class), utilizing holdout testing sets, and then, based on \cite{Kemker:2018}, used it to calculate the normalized average accuracy over the whole sequence:
\begin{equation}
    \Omega_{all} = \frac{1}{T}\sum_{t=1}^{T}\frac{\alpha_t}{\alpha_{\textrm{offline},t}},
\end{equation}

\noindent where $\alpha_t$ is the model performance after $t$ classes and $\alpha_{\textrm{offline},t}$ is the optimal performance obtained by the offline learner.

To make our scenario more challenging, we assumed that we did not know the classes of the original set, only the recommendation labels. This allowed us to create a complex decision space without explicit knowledge of its subspaces and with a lot of potential for local concept drifts.

\subsection{Results}
\noindent \textbf{Performance on stationary continual learning.} Firstly, we evaluated the performance of RSB against the reference approaches in class-incremental continual learning with stationary properties. That means there was no concept drift present in the data and the main challenge lay in aggregating learned knowledge and avoiding catastrophic forgetting. We used this scenario first as an ablation study, to show that RSB is capable of learning newly arriving classes, without forgetting the previously seen ones. Tab.~\ref{tab:stat} shows the normalized average accuracy results over the five used benchmarks, while Fig.~\ref{fig:stat1} depicts the changes in accuracy over time, calculated after each class (subconcept) batch. We omit the MNIST plot as it has identical characteristics as the FASHION plot.

In the presented results, we can see that all of the considered experience replay approaches were able to obtain satisfactory performance on the stationary sequences, slightly below the offline upper bound. They significantly improved upon the naive fine-tuning (NN), which severely suffered from catastrophic forgetting. The simple class buffers performed similarly on average. Holding instances of the earliest classes (CB0) turned out to be a bit better approach on simpler benchmarks, while giving a higher priority to the newer instances (CB1) resulted in higher accuracy on CIFAR10 and IMAGENET10. The more sophisticated centroid-driven experience replay (SB, RSB) provided even higher quality on all sequences by maintaining more diversified memory buffers per recommendation class. Finally, the results indicate that our method is often capable of improving upon the simpler centroid-based method (SB), most likely by correcting partially inaccurate clusters.

\begin{table}[h]
	\caption{The normalized average accuracy (absolute values for the offline baseline) for stationary sequences. }
	\vspace*{-0.1cm}
	\centering
	\scalebox{0.8}{
		\begin{tabular}[H]{l >{\centering\arraybackslash} m{1.2cm} >{\centering\arraybackslash} m{1.2cm} >{\centering\arraybackslash} m{1.2cm} >{\centering\arraybackslash} m{1.2cm} >{\centering\arraybackslash} m{1.2cm}}
			\toprule	
			\textit{Model} & MNIST & FASH & SVHN & CIF10 & IMGN10\\
			\midrule
			OFFLINE & 1.0 & 0.9865 & 1.0 & 1.0 & 1.0\\
			NN & 0.5529 & 0.5603 & 0.5529 & 0.5596 & 0.4886\\
			ER-CB0 & 0.9537 & 0.9554 & 0.9414 & 0.9106 & 0.8828\\
			ER-CB1 & 0.8754 & 0.8990 & 0.9235 & 0.9298 & 0.9349\\
			\midrule
			ER-SB & 0.9897 & 0.9739 & 0.9750 & 0.9675 & \textbf{0.9513}\\
			ER-RSB & \textbf{0.9967} & \textbf{0.9926} & \textbf{0.9816} & \textbf{0.9740} & 0.9405\\
			\bottomrule
		\end{tabular}
	}
	\label{tab:stat}
\end{table}

\begin{table}[h]
	\caption{The normalized average accuracy (absolute values for the offline baseline) for drifting sequences.}
	\vspace*{-0.1cm}
	\centering
	\scalebox{0.8}{
		\begin{tabular}[H]{l >{\centering\arraybackslash} m{1.2cm} >{\centering\arraybackslash} m{1.2cm} >{\centering\arraybackslash} m{1.2cm} >{\centering\arraybackslash} m{1.2cm} >{\centering\arraybackslash} m{1.2cm}}
			\toprule	
			\textit{Model} & MNIST & FASH & SVHN & CIF10 & IMGN10\\
			\midrule
			OFFLINE & 1.0 & 0.9744 & 1.0 & 1.0 & 1.0\\
			NN & 0.5894 & 0.6043 & 0.5872 & 0.5884 & 0.4546\\
			ER-CB0 & 0.5977 & 0.6473 & 0.5494 & 0.5635 & 0.6084\\
			ER-CB1 & 0.7422 & 0.7931 & 0.7743 & 0.7918 & 0.8540\\
			\midrule
			ER-SB & 0.7268 & 0.7341 & 0.7267 & 0.7004 & 0.6696\\
			ER-RSB & \textbf{0.9938} & \textbf{0.9745} & \textbf{0.9722} & \textbf{0.9545} & \textbf{0.9187}\\
			\bottomrule
		\end{tabular}
	}
	\label{tab:drift}
\end{table}

In Fig. \ref{fig:stat1} we can clearly see that RSB displayed stable incremental learning capabilities and was not affected by catastrophic forgetting. This is especially visible on FASHION, SVHN and CIFAR10 data sets, where with the increasing number of classes reference methods displayed drops of performance, while RSB achieved stable results for all arriving classes. For SVHN, we can see that CB0 returned to similar performance as RSB after the 8-th class -- but the intermediate learning process between classes no. 4 and 8 was significantly impaired. SB was much more resilient to forgetting, yet it performed slightly worse than RSB on 3 out of 4 sequences and on average. This allows us to conclude that RSB is robust to both catastrophic forgetting and false concept drift detection on stationary data.

\begin{figure*}[!h]
	\centering
	\leftskip70pt\begin{subfigure}{.8\columnwidth}
		\begin{tikzpicture}
		
		\begin{axis}
		[
		title={FASHION (DRIFT)},
		title style={yshift=-1.5ex},
		width=1\textwidth,
		height=.7\textwidth,
		ymajorgrids=true,
		ylabel near ticks,
		y tick label style={
			font=\scriptsize,
			/pgf/number format/.cd,
			fixed,
			fixed zerofill,
			precision=1,
			/tikz/.cd
		},
		ylabel style={font=\footnotesize, at={(-0.15,0.5)}},
		extra y ticks=0.5,
		extra y tick labels={},
		extra y tick style={
			ymajorgrids=true,
			ytick style={
				/pgfplots/major tick length=0pt,
			},
			grid style={
				gray,
				dashed,
				/pgfplots/on layer=axis foreground,
			},
		},
		grid style={dashed,white!90!black},
		xtick style={draw=none},
		xtick=data,
		xticklabels={0,,2,,4,,6,,8,,10,,12,,14,,16,,18,,20,,22,,24,,26,,28,,},
		xlabel style={font=\footnotesize},
		x tick label style={font=\scriptsize},
%		legend image post style={scale=0.7},
		legend style={font=\tiny,column sep=1ex, legend columns=1, at={(0.65,0.4)}, anchor=north west, draw opacity=0.8},
		legend entries={RSB,SB,CB1},
		]
		
		\node[font=\scriptsize] at (axis cs: 0.5,0.54) {$Rand$};
		
		\addplot [	
		draw=red2,mark=*,mark options={red2,scale=0.3},line width=1.2pt
		] coordinates {(0, 1.0) (1, 0.9930000305175781) (2, 0.9953333735466003) (3, 0.9759999513626099) (4, 0.9729999899864197) (5, 0.965499997138977) (6, 0.9539999961853027) (7, 0.9635000228881836) (8, 0.9032856822013855) (9, 0.8709999322891235) (10, 0.9004285931587219) (11, 0.9160000085830688) (12, 0.918222188949585) (13, 0.9238999485969543) (14, 0.9236000180244446) (15, 0.9161999821662903) (16, 0.9026999473571777) (17, 0.9078999757766724) (18, 0.9048999547958374) (19, 0.9621999859809875) (20, 0.962399959564209) (21, 0.9613999128341675) (22, 0.9690999984741211) (23, 0.9687999486923218) (24, 0.9683000445365906) (25, 0.9761000871658325) (26, 0.9815000295639038) (27, 0.9819999933242798) (28, 0.9811000823974609) (29, 0.9785000085830688)};
		
		\addplot [	
		draw=blue2,mark=square,mark options={blue2,scale=0.3},line width=1pt
		] coordinates {(0, 1.0) (1, 0.9959999918937683) (2, 0.9953333735466003) (3, 0.9789999723434448) (4, 0.7922499775886536) (5, 0.3375000059604645) (6, 0.4772000312805176) (7, 0.5460000038146973) (8, 0.5870000123977661) (9, 0.5434285402297974) (10, 0.6607142686843872) (11, 0.6777499914169312) (12, 0.6970000267028809) (13, 0.7246000170707703) (14, 0.7233999967575073) (15, 0.6721999645233154) (16, 0.6931999325752258) (17, 0.6575999855995178) (18, 0.6972000598907471) (19, 0.7298000454902649) (20, 0.7183000445365906) (21, 0.6687000393867493) (22, 0.7315999269485474) (23, 0.6987999677658081) (24, 0.6929999589920044) (25, 0.720300018787384) (26, 0.7731999754905701) (27, 0.7776000499725342) (28, 0.7639999389648438) (29, 0.7254999876022339)};
		
		\addplot [	
		draw=brown1,mark=triangle,mark options={brown1,scale=0.3},line width=1pt
		] coordinates {(0, 1.0) (1, 0.9954999685287476) (2, 0.9950000643730164) (3, 0.9449999928474426) (4, 0.9679999947547913) (5, 0.6170000433921814) (6, 0.8159999847412109) (7, 0.8218333721160889) (8, 0.715571403503418) (9, 0.5049999952316284) (10, 0.6268571615219116) (11, 0.8252500295639038) (12, 0.8117778301239014) (13, 0.850100040435791) (14, 0.6554999947547913) (15, 0.41130000352859497) (16, 0.40890002250671387) (17, 0.5765999555587769) (18, 0.6112999320030212) (19, 0.7062000036239624) (20, 0.6904000043869019) (21, 0.7904999852180481) (22, 0.7820000052452087) (23, 0.8515999913215637) (24, 0.7524999380111694) (25, 0.8222000002861023) (26, 0.7911999225616455) (27, 0.953299880027771) (28, 0.9262000322341919) (29, 0.9628999829292297)};
		
		\end{axis}
		
		\end{tikzpicture}
		\end{subfigure}%
		\hspace*{-0.6cm}
		\begin{subfigure}{.8\columnwidth}
		\begin{tikzpicture}
		
		\begin{axis}
		[
		title={SVHN (DRIFT)},
		title style={yshift=-1.5ex},
		width=1\textwidth,
		height=.7\textwidth,
		ymajorgrids=true,
		ylabel near ticks,
		y tick label style={
			font=\scriptsize,
			/pgf/number format/.cd,
			fixed,
			fixed zerofill,
			precision=1,
			/tikz/.cd
		},
		ylabel style={font=\footnotesize, at={(-0.15,0.5)}},
		extra y ticks=0.5,
		extra y tick labels={},
		extra y tick style={
			ymajorgrids=true,
			ytick style={
				/pgfplots/major tick length=0pt,
			},
			grid style={
				gray,
				dashed,
				/pgfplots/on layer=axis foreground,
			},
		},
		grid style={dashed,white!90!black},
		xtick style={draw=none},
		xtick=data,
		xticklabels={0,,2,,4,,6,,8,,10,,12,,14,,16,,18,,20,,22,,24,,26,,28,,},
		xlabel style={font=\footnotesize},
		x tick label style={font=\scriptsize}
		]
		
		\node[font=\scriptsize] at (axis cs: 0.5,0.54) {$Rand$};
		
		\addplot [	
		draw=red2,mark=*,mark options={red2,scale=0.3},line width=1.2pt
		] coordinates {(0, 0.9862385392189026) (1, 0.9906378984451294) (2, 0.9874549508094788) (3, 0.9834433197975159) (4, 0.9878103137016296) (5, 0.9824420213699341) (6, 0.982639491558075) (7, 0.9841817021369934) (8, 0.9730514883995056) (9, 0.9261530637741089) (10, 0.8944324254989624) (11, 0.9662713408470154) (12, 0.9666422605514526) (13, 0.9669049382209778) (14, 0.9460415840148926) (15, 0.9765939712524414) (16, 0.9770761728286743) (17, 0.9766756296157837) (18, 0.9760730862617493) (19, 0.9735269546508789) (20, 0.9751654863357544) (21, 0.9746795892715454) (22, 0.9752613306045532) (23, 0.975528359413147) (24, 0.9787886738777161) (25, 0.976584792137146) (26, 0.9762333035469055) (27, 0.9768341779708862) (28, 0.9764999151229858) (29, 0.9770940542221069)};
		
		\addplot [	
		draw=blue2,mark=square,mark options={blue2,scale=0.3},line width=1pt
		] coordinates {(0, 0.9856651425361633) (1, 0.9899814128875732) (2, 0.9880298972129822) (3, 0.9825330972671509) (4, 0.755199670791626) (5, 0.3285720646381378) (6, 0.4311158061027527) (7, 0.5237798690795898) (8, 0.5882360339164734) (9, 0.7348189949989319) (10, 0.7220228910446167) (11, 0.7291600108146667) (12, 0.7619836926460266) (13, 0.7809355854988098) (14, 0.7876583337783813) (15, 0.7899948954582214) (16, 0.8064007759094238) (17, 0.7699225544929504) (18, 0.747373104095459) (19, 0.7211901545524597) (20, 0.608006477355957) (21, 0.7064728736877441) (22, 0.7193835377693176) (23, 0.7536823749542236) (24, 0.6431561708450317) (25, 0.6897427439689636) (26, 0.6837598085403442) (27, 0.6775504350662231) (28, 0.6528235673904419) (29, 0.7433116436004639)};
		
		\addplot [	
		draw=brown1,mark=triangle,mark options={brown1,scale=0.3},line width=1pt
		] coordinates {(0, 0.9868119359016418) (1, 0.989679753780365) (2, 0.774684727191925) (3, 0.9779305458068848) (4, 0.9822330474853516) (5, 0.7717874050140381) (6, 0.7992492914199829) (7, 0.6649625301361084) (8, 0.45886150002479553) (9, 0.7103622555732727) (10, 0.7673071622848511) (11, 0.7965478897094727) (12, 0.8150430917739868) (13, 0.7967137694358826) (14, 0.8168123364448547) (15, 0.6257766485214233) (16, 0.38146770000457764) (17, 0.5957722067832947) (18, 0.8573354482650757) (19, 0.6321672201156616) (20, 0.6856555342674255) (21, 0.6663317680358887) (22, 0.7890208959579468) (23, 0.7692780494689941) (24, 0.6948086023330688) (25, 0.8746917843818665) (26, 0.7956507802009583) (27, 0.9126065969467163) (28, 0.8860828280448914) (29, 0.9524322748184204)};
		
		\end{axis}
		
		\end{tikzpicture}%
	\end{subfigure}\\
	\vspace{0.1cm}
	\leftskip70pt\begin{subfigure}{.8\columnwidth}
		\begin{tikzpicture}
		
		\begin{axis}
		[
		title={CIFAR10 (DRIFT)},
		title style={yshift=-1.5ex},
		width=1\textwidth,
		height=.7\textwidth,
		ymajorgrids=true,
		ylabel near ticks,
		y tick label style={
			font=\scriptsize,
			/pgf/number format/.cd,
			fixed,
			fixed zerofill,
			precision=1,
			/tikz/.cd
		},
		ylabel style={font=\footnotesize, at={(-0.15,0.5)}},
		extra y ticks=0.5,
		extra y tick labels={},
		extra y tick style={
			ymajorgrids=true,
			ytick style={
				/pgfplots/major tick length=0pt,
			},
			grid style={
				gray,
				dashed,
				/pgfplots/on layer=axis foreground,
			},
		},
		grid style={dashed,white!90!black},
		xtick style={draw=none},
		xtick=data,
		xticklabels={0,,2,,4,,6,,8,,10,,12,,14,,16,,18,,20,,22,,24,,26,,28,,},
		xlabel style={font=\footnotesize},
		x tick label style={font=\scriptsize}
		]
		
		\node[font=\scriptsize] at (axis cs: 0.5,0.54) {$Rand$};
		
		\addplot [	
		draw=red2,mark=*,mark options={red2,scale=0.3},line width=1.2pt
		] coordinates {(0, 0.9919999837875366) (1, 0.9930000305175781) (2, 0.9943333268165588) (3, 0.9607499837875366) (4, 0.9647499918937683) (5, 0.950249969959259) (6, 0.9406000375747681) (7, 0.9611666202545166) (8, 0.9609999656677246) (9, 0.9639999270439148) (10, 0.9430000185966492) (11, 0.9468749761581421) (12, 0.945555567741394) (13, 0.9441999197006226) (14, 0.9430000185966492) (15, 0.8748000264167786) (16, 0.9156000018119812) (17, 0.9210000038146973) (18, 0.9413000345230103) (19, 0.953700065612793) (20, 0.9521999359130859) (21, 0.9522000551223755) (22, 0.9587000012397766) (23, 0.9587000012397766) (24, 0.9660000801086426) (25, 0.9668000340461731) (26, 0.9673999547958374) (27, 0.9675000309944153) (28, 0.9677000045776367) (29, 0.9668000340461731)};
		
		\addplot [	
		draw=blue2,mark=square,mark options={blue2,scale=0.3},line width=1pt
		] coordinates {(0, 0.9929999709129333) (1, 0.9945000410079956) (2, 0.9953333735466003) (3, 0.9614999890327454) (4, 0.7354999780654907) (5, 0.5) (6, 0.6060000061988831) (7, 0.6700000166893005) (8, 0.711142897605896) (9, 0.7070000767707825) (10, 0.6811428070068359) (11, 0.627625048160553) (12, 0.6667777299880981) (13, 0.6829000115394592) (14, 0.6256999969482422) (15, 0.6574999690055847) (16, 0.6534000039100647) (17, 0.7148000001907349) (18, 0.6362999677658081) (19, 0.6166999936103821) (20, 0.7303999662399292) (21, 0.7297999262809753) (22, 0.7338000535964966) (23, 0.735200047492981) (24, 0.6502000093460083) (25, 0.6282000541687012) (26, 0.6343000531196594) (27, 0.5770000219345093) (28, 0.5807000398635864) (29, 0.5758999586105347)};
		
		\addplot [	
		draw=brown1,mark=triangle,mark options={brown1,scale=0.3},line width=1pt
		] coordinates {(0, 0.9959999918937683) (1, 0.9950000047683716) (2, 0.9950000643730164) (3, 0.9597499966621399) (4, 0.9612500667572021) (5, 0.7447500228881836) (6, 0.7803999781608582) (7, 0.7843332886695862) (8, 0.8440000414848328) (9, 0.8822856545448303) (10, 0.5994285941123962) (11, 0.8416249752044678) (12, 0.676111102104187) (13, 0.6873999834060669) (14, 0.675000011920929) (15, 0.381600022315979) (16, 0.6410999894142151) (17, 0.8746000528335571) (18, 0.7858999371528625) (19, 0.849000096321106) (20, 0.774399995803833) (21, 0.6660999655723572) (22, 0.8447999954223633) (23, 0.8391000032424927) (24, 0.6000999808311462) (25, 0.694599986076355) (26, 0.6924999952316284) (27, 0.759600043296814) (28, 0.960800051689148) (29, 0.9668000340461731)};
		
		\end{axis}
		
		\end{tikzpicture}
		\end{subfigure}%
		\hspace*{-0.6cm}
		\begin{subfigure}{.8\columnwidth}
		\begin{tikzpicture}
		
		\begin{axis}
		[
		title={IMAGENET10 (DRIFT)},
		title style={yshift=-1.5ex},
		width=1\textwidth,
		height=.7\textwidth,
		ymajorgrids=true,
		ylabel near ticks,
		y tick label style={
			font=\scriptsize,
			/pgf/number format/.cd,
			fixed,
			fixed zerofill,
			precision=1,
			/tikz/.cd
		},
		ylabel style={font=\footnotesize, at={(-0.15,0.5)}},
		extra y ticks=0.5,
		extra y tick labels={},
		extra y tick style={
			ymajorgrids=true,
			ytick style={
				/pgfplots/major tick length=0pt,
			},
			grid style={
				gray,
				dashed,
				/pgfplots/on layer=axis foreground,
			},
		},
		grid style={dashed,white!90!black},
		xtick style={draw=none},
		xtick=data,
		xticklabels={0,,2,,4,,6,,8,,10,,12,,14,,16,,18,,20,,22,,24,,26,,28,,},
		xlabel style={font=\footnotesize},
		x tick label style={font=\scriptsize}
		]
		
		\node[font=\scriptsize] at (axis cs: 0.5,0.54) {$Rand$};
		
		\addplot [	
		draw=red2,mark=*,mark options={red2,scale=0.3},line width=1.2pt
		] coordinates {(0, 0.9399999976158142) (1, 0.9399999976158142) (2, 0.9533333778381348) (3, 0.9450000524520874) (4, 0.9549999833106995) (5, 0.9650000333786011) (6, 0.9480000734329224) (7, 0.9433333873748779) (8, 0.9428572058677673) (9, 0.9314286112785339) (10, 0.9028571844100952) (11, 0.8725000619888306) (12, 0.8844444751739502) (13, 0.8700000047683716) (14, 0.9120000004768372) (15, 0.9180000424385071) (16, 0.9239999651908875) (17, 0.9279999732971191) (18, 0.9259999990463257) (19, 0.9239999651908875) (20, 0.9140000343322754) (21, 0.8500000238418579) (22, 0.8480000495910645) (23, 0.8920000195503235) (24, 0.9100000262260437) (25, 0.9200000762939453) (26, 0.9220000505447388) (27, 0.9239999651908875) (28, 0.9239999651908875) (29, 0.9300000071525574)};
		
		\addplot [	
		draw=blue2,mark=square,mark options={blue2,scale=0.3},line width=1pt
		] coordinates {(0, 1.0) (1, 0.9600000381469727) (2, 0.9533333778381348) (3, 0.9449999928474426) (4, 0.75) (5, 0.5249999761581421) (6, 0.6079999804496765) (7, 0.6466666460037231) (8, 0.699999988079071) (9, 0.6800000071525574) (10, 0.5771428346633911) (11, 0.4650000035762787) (12, 0.5311110615730286) (13, 0.49799999594688416) (14, 0.593999981880188) (15, 0.5839999914169312) (16, 0.6039999723434448) (17, 0.593999981880188) (18, 0.593999981880188) (19, 0.6840000152587891) (20, 0.5600000023841858) (21, 0.5160000324249268) (22, 0.6180000305175781) (23, 0.7440000772476196) (24, 0.7300000190734863) (25, 0.6480000019073486) (26, 0.6539999842643738) (27, 0.6380000114440918) (28, 0.6999999284744263) (29, 0.7860000133514404)};
		
		\addplot [	
		draw=brown1,mark=triangle,mark options={brown1,scale=0.3},line width=1pt
		] coordinates {(0, 1.0) (1, 0.9800000190734863) (2, 0.9600000381469727) (3, 0.9300000667572021) (4, 0.9650000333786011) (5, 0.9700000286102295) (6, 0.9319999814033508) (7, 0.9433333873748779) (8, 0.9428572058677673) (9, 0.9428572058677673) (10, 0.8057142496109009) (11, 0.8325000405311584) (12, 0.8866667151451111) (13, 0.7640000581741333) (14, 0.8159999847412109) (15, 0.6340000033378601) (16, 0.8899999856948853) (17, 0.8300000429153442) (18, 0.8980000615119934) (19, 0.8999999761581421) (20, 0.718000054359436) (21, 0.8560000658035278) (22, 0.8920000195503235) (23, 0.8659999966621399) (24, 0.39000001549720764) (25, 0.5559999942779541) (26, 0.8119999766349792) (27, 0.8999999761581421) (28, 0.9300000071525574) (29, 0.8780000805854797)};
		
		\end{axis}
		
		\end{tikzpicture}%
	\end{subfigure}%
	%\vspace*{-2mm}
	\caption{Average accuracy over all classes for drifting class-incremental sequences. Drifts occur in batches 4, 5, 9, 10, 14, 15, 19, 20, 24 and 25.}
	\label{fig:drift1}
\end{figure*}
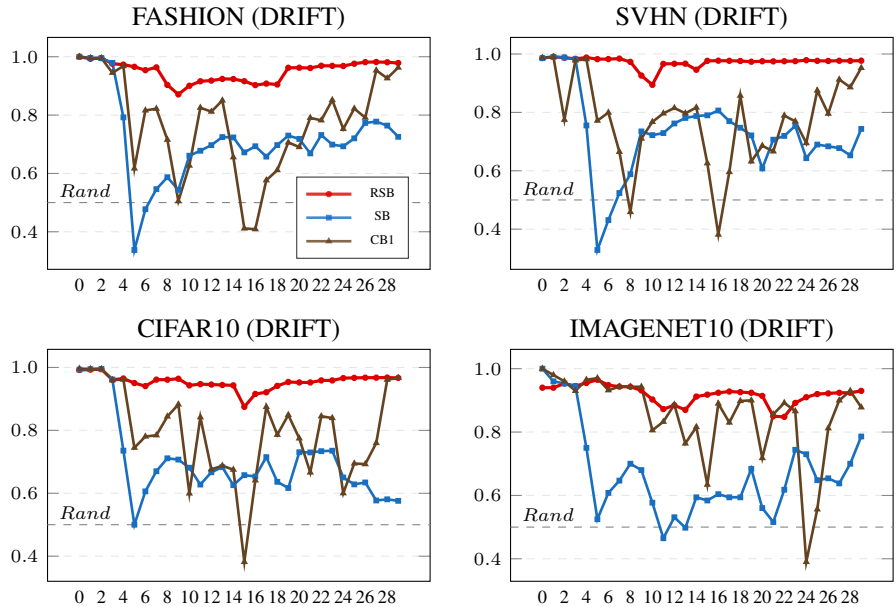

\smallskip
\noindent \textbf{Performance on continual learning under concept drift.} After establishing that RSB displays robustness to catastrophic forgetting, we needed to evaluate its capability of simultaneous incremental learning and adaptation to drift. We used the same five benchmarks that now were injected with concept drift as discussed in Sec. 5.1. This way we should be able to see if RSB is able to detect changes on previously learned classes and correctly modify the underlying classifier to update its stored knowledge. Tab.~\ref{tab:drift} shows the normalized average accuracy results over five used benchmarks, while Fig.~\ref{fig:drift1} depicts the changes in accuracy over time. Again, we omit MNIST plot as it has identical characteristics as the FASHION plot.

\begin{figure*}[!h]
	\centering
	\leftskip70pt\begin{subfigure}{.8\columnwidth}
		\begin{tikzpicture}
		
		\begin{axis}
		[
		title={FASHION (STAT)},
		title style={yshift=-1.5ex},
		width=1\textwidth,
		height=.7\textwidth,
		ymin=0.25,
		ymajorgrids=true,
		ylabel near ticks,
		y tick label style={
			font=\scriptsize,
			/pgf/number format/.cd,
			fixed,
			fixed zerofill,
			precision=1,
			/tikz/.cd
		},
		ylabel style={font=\footnotesize, at={(-0.15,0.5)}},
		extra y ticks=0.5,
		extra y tick labels={},
		extra y tick style={
			ymajorgrids=true,
			ytick style={
				/pgfplots/major tick length=0pt,
			},
			grid style={
				gray,
				dashed,
				/pgfplots/on layer=axis foreground,
			},
		},
		grid style={dashed,white!90!black},
		xtick style={draw=none},
		xtick=data,
		xlabel style={font=\footnotesize},
		x tick label style={font=\scriptsize},
		%		legend image post style={scale=0.7},
		legend style={font=\tiny,column sep=1ex, legend columns=-1, at={(0.08,0.2)}, anchor=north west, draw opacity=0.8},
		legend entries={RSB,SB,CB0},
		]
		
		\node[font=\scriptsize] at (axis cs: 0.5,0.55) {$Rand$};
		
		\addplot [	
		draw=red2,mark=*,mark options={red2,scale=0.3},line width=1.2pt
		] coordinates {(0, 1.0) (1, 0.9939999580383301) (2, 0.9953333735466003) (3, 0.965749979019165) (4, 0.9668000340461731) (5, 0.9718334078788757) (6, 0.9688571691513062) (7, 0.9740000367164612) (8, 0.9760000705718994) (9, 0.9793999791145325)};
		
		\addplot [	
		draw=blue2,mark=square,mark options={blue2,scale=0.3},line width=1pt
		] coordinates {(0, 1.0) (1, 0.9939999580383301) (2, 0.9956666827201843) (3, 0.9787499904632568) (4, 0.9687999486923218) (5, 0.9503333568572998) (6, 0.926428496837616) (7, 0.9441250562667847) (8, 0.913777768611908) (9, 0.9161000061035156)};
		
		\addplot [	
		draw=brown1,mark=triangle,mark options={brown1,scale=0.3},line width=1pt
		] coordinates {(0, 1.0) (1, 0.9950000047683716) (2, 0.9950000643730164) (3, 0.9735000133514404) (4, 0.9312000274658203) (5, 0.8829999566078186) (6, 0.920714259147644) (7, 0.9399999976158142) (8, 0.8436667323112488) (9, 0.9028000450134277)};
		
		\end{axis}
		
		\end{tikzpicture}
		\end{subfigure}%
		\hspace*{-0.6cm}
		\begin{subfigure}{.8\columnwidth}
		\begin{tikzpicture}
		
		\begin{axis}
		[
		title={SVHN (STAT)},
		title style={yshift=-1.5ex},
		width=1\textwidth,
		height=.7\textwidth,
		ymin=0.35,
		ymajorgrids=true,
		ylabel near ticks,
		y tick label style={
			font=\scriptsize,
			/pgf/number format/.cd,
			fixed,
			fixed zerofill,
			precision=1,
			/tikz/.cd
		},
		ylabel style={font=\footnotesize, at={(-0.15,0.5)}},
		extra y ticks=0.5,
		extra y tick labels={},
		extra y tick style={
			ymajorgrids=true,
			ytick style={
				/pgfplots/major tick length=0pt,
			},
			grid style={
				gray,
				dashed,
				/pgfplots/on layer=axis foreground,
			},
		},
		grid style={dashed,white!90!black},
		xtick style={draw=none},
		xtick=data,
		xlabel style={font=\footnotesize},
		x tick label style={font=\scriptsize}
		]
		
		\node[font=\scriptsize] at (axis cs: 0.5,0.54) {$Rand$};
		
		\addplot [	
		draw=red2,mark=*,mark options={red2,scale=0.3},line width=1.2pt
		] coordinates {(0, 0.9891055226325989) (1, 0.9899590015411377) (2, 0.9883856773376465) (3, 0.9827327728271484) (4, 0.9785023927688599) (5, 0.9814836382865906) (6, 0.9795778393745422) (7, 0.9774003028869629) (8, 0.974879801273346) (9, 0.9742718935012817)};
		
		\addplot [	
		draw=blue2,mark=square,mark options={blue2,scale=0.3},line width=1pt
		] coordinates {(0, 0.9885321259498596) (1, 0.9903212785720825) (2, 0.9880134463310242) (3, 0.9833261966705322) (4, 0.9803832769393921) (5, 0.9797250628471375) (6, 0.9787276983261108) (7, 0.9440895318984985) (8, 0.9584038853645325) (9, 0.9582546949386597)};
		
		\addplot [	
		draw=brown1,mark=triangle,mark options={brown1,scale=0.3},line width=1pt
		] coordinates {(0, 0.9873853325843811) (1, 0.9901700019836426) (2, 0.9861245155334473) (3, 0.9708682894706726) (4, 0.7855607271194458) (5, 0.9705799221992493) (6, 0.9300169348716736) (7, 0.8674534559249878) (8, 0.9746512174606323) (9, 0.9509387016296387)};
		
		\end{axis}
		
		\end{tikzpicture}%
	\end{subfigure}\\
	\vspace{0.1cm}
	\leftskip70pt\begin{subfigure}{.8\columnwidth}
		\begin{tikzpicture}
		
		\begin{axis}
		[
		title={CIFAR10 (STAT)},
		title style={yshift=-1.5ex},
		width=1\textwidth,
		height=.7\textwidth,
		ymin=0.35,
		ymajorgrids=true,
		ylabel near ticks,
		y tick label style={
			font=\scriptsize,
			/pgf/number format/.cd,
			fixed,
			fixed zerofill,
			precision=1,
			/tikz/.cd
		},
		ylabel style={font=\footnotesize, at={(-0.15,0.5)}},
		extra y ticks=0.5,
		extra y tick labels={},
		extra y tick style={
			ymajorgrids=true,
			ytick style={
				/pgfplots/major tick length=0pt,
			},
			grid style={
				gray,
				dashed,
				/pgfplots/on layer=axis foreground,
			},
		},
		grid style={dashed,white!90!black},
		xtick style={draw=none},
		xtick=data,
		xlabel style={font=\footnotesize},
		x tick label style={font=\scriptsize}
		]
		
		\node[font=\scriptsize] at (axis cs: 0.5,0.54) {$Rand$};
		
		\addplot [	
		draw=red2,mark=*,mark options={red2,scale=0.3},line width=1.2pt
		] coordinates {(0, 0.9929999709129333) (1, 0.9930000305175781) (2, 0.9950000643730164) (3, 0.9637500643730164) (4, 0.9583999514579773) (5, 0.9654999375343323) (6, 0.9684286117553711) (7, 0.9675000309944153) (8, 0.9671111702919006) (9, 0.968000054359436)};
		
		\addplot [	
		draw=blue2,mark=square,mark options={blue2,scale=0.3},line width=1pt
		] coordinates {(0, 0.9950000047683716) (1, 0.9930000305175781) (2, 0.9956666827201843) (3, 0.9619999527931213) (4, 0.954800009727478) (5, 0.9350000023841858) (6, 0.9625714421272278) (7, 0.9595000147819519) (8, 0.9592222571372986) (9, 0.9578999280929565)};
		
		\addplot [	
		draw=brown1,mark=triangle,mark options={brown1,scale=0.3},line width=1pt
		] coordinates {(0, 0.9959999918937683) (1, 0.9919999837875366) (2, 0.9946666359901428) (3, 0.734749972820282) (4, 0.7918000221252441) (5, 0.9358332753181458) (6, 0.9212856888771057) (7, 0.9318749308586121) (8, 0.9324444532394409) (9, 0.8755000233650208)};
		
		\end{axis}
		
		\end{tikzpicture}
		\end{subfigure}%
		\hspace*{-0.6cm}
		\begin{subfigure}{.8\columnwidth}
		\begin{tikzpicture}
		
		\begin{axis}
		[
		title={IMAGENET10 (STAT)},
		title style={yshift=-1.5ex},
		width=1\textwidth,
		height=.7\textwidth,
		ymin=0.35,
		ymajorgrids=true,
		ylabel near ticks,
		y tick label style={
			font=\scriptsize,
			/pgf/number format/.cd,
			fixed,
			fixed zerofill,
			precision=1,
			/tikz/.cd
		},
		ylabel style={font=\footnotesize, at={(-0.15,0.5)}},
		extra y ticks=0.5,
		extra y tick labels={},
		extra y tick style={
			ymajorgrids=true,
			ytick style={
				/pgfplots/major tick length=0pt,
			},
			grid style={
				gray,
				dashed,
				/pgfplots/on layer=axis foreground,
			},
		},
		grid style={dashed,white!90!black},
		xtick style={draw=none},
		xtick=data,
		xlabel style={font=\footnotesize},
		x tick label style={font=\scriptsize}
		]
		
		\node[font=\scriptsize] at (axis cs: 0.5,0.54) {$Rand$};
		
		\addplot [	
		draw=red2,mark=*,mark options={red2,scale=0.3},line width=1.2pt
		] coordinates {(0, 0.9420000166893005) (1, 0.9800000190734863) (2, 0.9533333778381348) (3, 0.9390000619888306) (4, 0.9440000653266907) (5, 0.9399999976158142) (6, 0.9371428489685059) (7, 0.9275000095367432) (8, 0.9377778172492981) (9, 0.9299999475479126)};
		
		\addplot [	
		draw=blue2,mark=square,mark options={blue2,scale=0.3},line width=1pt
		] coordinates {(0, 0.9900177641435377) (1, 0.9800000190734863) (2, 0.9600000381469727) (3, 0.9500000476837158) (4, 0.9479999542236328) (5, 0.9399999976158142) (6, 0.9371428489685059) (7, 0.9275000095367432) (8, 0.9399999976158142) (9, 0.9299999475479126)};
		
		\addplot [	
		draw=brown1,mark=triangle,mark options={brown1,scale=0.3},line width=1pt
		] coordinates {(0, 1.0) (1, 1.0) (2, 0.9600000381469727) (3, 0.9650000333786011) (4, 0.9360000491142273) (5, 0.7899999618530273) (6, 0.925714373588562) (7, 0.7575000524520874) (8, 0.7200000286102295) (9, 0.7740000486373901)};
		
		\end{axis}
		
		\end{tikzpicture}%
	\end{subfigure}%
	%\vspace*{-2mm}
	\caption{Average accuracy over all classes for stationary class-incremental sequences.}
	\label{fig:stat1}
\end{figure*}
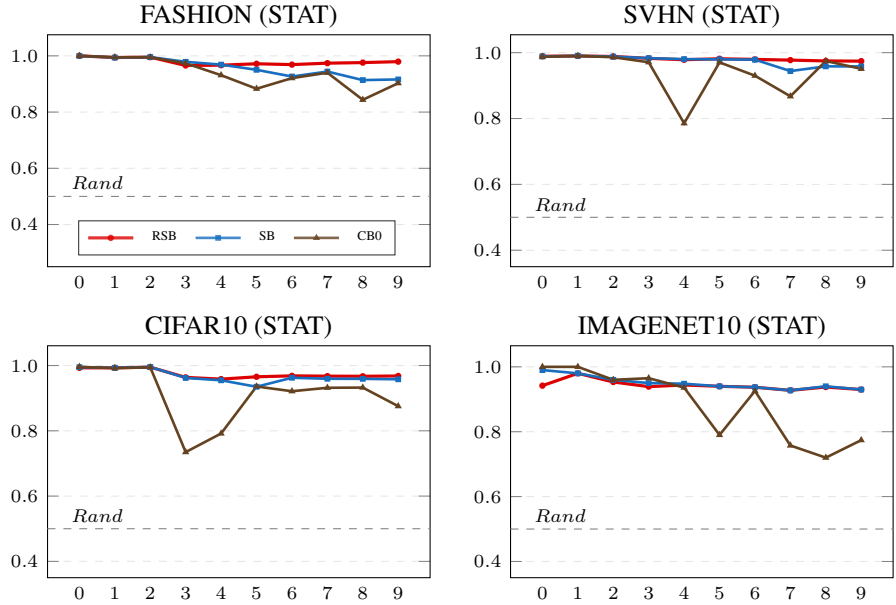

For continual learning under concept drift we can see significant differences among the examined algorithms. Neither CB0, CB1 or SB were capable of keeping up with the presence of concept drift in the data. The main reason for that was the fact that CB0 and SB kept outdated instances in their buffers, impeding the adaptation process by forcing the model to retain obsolete concepts. On the other hand, CB1 adapted to newer concepts much better than CB0, but it was not able to store instances for older classes, which inevitably led to catastrophic forgetting. For all five data sets the proposed RSB displayed the most stable performance, which is especially striking in the case of FASHION, SVHN and CIFAR10 sequences. By analyzing the plots we can see how the reference methods were significantly impacted by the first occurrence of concept drift, often dropping to similar or lower performance levels as the random approach. Sometimes they were slowly recovering their performance over time, but this was happening at an unacceptable rate.

\begin{figure}[h]
	\centering
	\leftskip30pt\begin{subfigure}{.8\columnwidth}
		\begin{tikzpicture}
		
		\begin{axis}
		[
		title={MNIST-C0},
		title style={yshift=-1.5ex},
		width=1\textwidth,
		height=.7\textwidth,
		ymajorgrids=true,
		ylabel near ticks,
		y tick label style={
			font=\scriptsize,
			/pgf/number format/.cd,
			fixed,
			fixed zerofill,
			precision=1,
			/tikz/.cd
		},
		ylabel style={font=\scriptsize, at={(-0.15,0.5)}},
		extra y ticks=0.5,
		extra y tick labels={},
		extra y tick style={
			ymajorgrids=true,
			ytick style={
				/pgfplots/major tick length=0pt,
			},
			grid style={
				gray,
				dashed,
				/pgfplots/on layer=axis foreground,
			},
		},
		grid style={dashed,white!90!black},
		xtick style={draw=none},
		xtick=data,
		xticklabels={0,,2,,4,,6,,8,,10,,12,,14,,16,,18,,20,,22,,24,,26,,28,,},
		xlabel style={font=\footnotesize},
		x tick label style={font=\scriptsize},
		%		legend image post style={scale=0.7},
%		legend style={font=\scriptsize,column sep=1ex, legend columns=-1, at={(0.1,0.96)}, anchor=north west, draw opacity=0.5},
%		legend entries={RSB,SB,CB1},
		]
		
		\node[font=\scriptsize] at (axis cs: 0.5,0.55) {$Rand$};
		
		\addplot [	
		draw=red2,mark=*,mark options={red2,scale=0.3},line width=1.2pt
		] coordinates {(0, 1.0) (1, 1.0) (2, 1.0) (3, 1.0) (4, 1.0) (5, 1.0) (6, 1.0) (7, 1.0) (8, 0.9979591965675354) (9, 0.9989795684814453) (10, 1.0) (11, 0.9989795684814453) (12, 0.9989795684814453) (13, 0.9989795684814453) (14, 0.9979591965675354) (15, 0.9969387650489807) (16, 0.9989795684814453) (17, 0.9989795684814453) (18, 0.9989795684814453) (19, 1.0) (20, 0.9979591965675354) (21, 0.9989795684814453) (22, 0.9979591965675354) (23, 0.9979591965675354) (24, 0.9989795684814453) (25, 0.9979591965675354) (26, 0.9989795684814453) (27, 0.9989795684814453) (28, 0.9989795684814453) (29, 0.9989795684814453)};
		
		\addplot [	
		draw=blue2,mark=square,mark options={blue2,scale=0.3},line width=1pt
		] coordinates {(0, 1.0) (1, 1.0) (2, 1.0) (3, 1.0) (4, 0.1755101978778839) (5, 0.11938775330781937) (6, 0.1418367326259613) (7, 0.12142857164144516) (8, 0.11632653325796127) (9, 0.09693877398967743) (10, 0.05612244829535484) (11, 0.015306122601032257) (12, 0.014285714365541935) (13, 0.019387755542993546) (14, 0.0367346927523613) (15, 0.025510204955935478) (16, 0.37755101919174194) (17, 0.31224489212036133) (18, 0.27244898676872253) (19, 0.1051020398736) (20, 0.06530611962080002) (21, 0.0693877562880516) (22, 0.06020408123731613) (23, 0.05306122452020645) (24, 0.02448979578912258) (25, 0.026530612260103226) (26, 0.5051020383834839) (27, 0.4979591965675354) (28, 0.44489794969558716) (29, 0.4285714328289032)};
		
		\addplot [	
		draw=brown1,mark=triangle,mark options={brown1,scale=0.3},line width=1pt
		] coordinates {(0, 1.0) (1, 1.0) (2, 1.0) (3, 1.0) (4, 1.0) (5, 1.0) (6, 1.0) (7, 1.0) (8, 0.9326530694961548) (9, 0.06326530873775482) (10, 0.052040815353393555) (11, 0.10306122153997421) (12, 0.052040815353393555) (13, 0.468367338180542) (14, 0.02448979578912258) (15, 0.0071428571827709675) (16, 1.0) (17, 0.9989795684814453) (18, 1.0) (19, 0.9918367266654968) (20, 0.9102040529251099) (21, 0.8693877458572388) (22, 0.9857142567634583) (23, 0.922448992729187) (24, 0.9969387650489807) (25, 0.11734694242477417) (26, 0.9989795684814453) (27, 0.9989795684814453) (28, 1.0) (29, 1.0)};
		
		%\draw [dashed] (4,0.0) -- (4,1.0);
		
		\end{axis}
		
		\end{tikzpicture}%
	\end{subfigure}\\
	\vspace{0.2cm}
	\begin{subfigure}{.8\columnwidth}
		\begin{tikzpicture}
		
		\begin{axis}
		[
		title={FASHION-C2},
		title style={yshift=-1.5ex},
		width=1\textwidth,
		height=.7\textwidth,
		ymajorgrids=true,
		ylabel near ticks,
		y tick label style={
			font=\scriptsize,
			/pgf/number format/.cd,
			fixed,
			fixed zerofill,
			precision=1,
			/tikz/.cd
		},
		ylabel style={font=\scriptsize, at={(-0.15,0.5)}},
		extra y ticks=0.5,
		extra y tick labels={},
		extra y tick style={
			ymajorgrids=true,
			ytick style={
				/pgfplots/major tick length=0pt,
			},
			grid style={
				gray,
				dashed,
				/pgfplots/on layer=axis foreground,
			},
		},
		grid style={dashed,white!90!black},
		xtick style={draw=none},
		xtick=data,
		xticklabels={2,,4,,6,,8,,10,,12,,14,,16,,18,,20,,22,,24,,26,,28,,},
		xlabel style={font=\footnotesize},
		x tick label style={font=\scriptsize},
				legend image post style={scale=0.7},
		legend style={font=\scriptsize,column sep=1ex, legend columns=1, at={(0.65,0.45)}, anchor=north west, draw opacity=0.8},
		legend entries={RSB,SB,CB1},
		]
		
		\node[font=\scriptsize] at (axis cs: 1.8,0.55) {$Rand$};
		
		\addplot [	
		draw=red2,mark=*,mark options={red2,scale=0.3},line width=1.2pt
		] coordinates {(2, 0.9980000257492065) (3, 0.9879999756813049) (4, 0.9509999752044678) (5, 0.9580000042915344) (6, 0.9649999737739563) (7, 0.9679999947547913) (8, 0.9919999837875366) (9, 0.9710000157356262) (10, 0.9570000171661377) (11, 0.9599999785423279) (12, 0.9549999833106995) (13, 0.9559999704360962) (14, 0.9879999756813049) (15, 0.9900000095367432) (16, 0.9940000176429749) (17, 0.9929999709129333) (18, 0.9959999918937683) (19, 0.9950000047683716) (20, 0.9909999966621399) (21, 0.9779999852180481) (22, 0.9890000224113464) (23, 0.9860000014305115) (24, 0.9829999804496765) (25, 0.9850000143051147) (26, 0.9919999837875366) (27, 0.9909999966621399) (28, 0.9940000176429749) (29, 0.9869999885559082)};
		
		\addplot [	
		draw=blue2,mark=square,mark options={blue2,scale=0.3},line width=1pt
		] coordinates {(2, 0.9990000128746033) (3, 0.9850000143051147) (4, 0.9779999852180481) (5, 0.9950000047683716) (6, 0.9929999709129333) (7, 0.9959999918937683) (8, 1.0) (9, 0.8880000114440918) (10, 0.8059999942779541) (11, 0.4569999873638153) (12, 0.28999999165534973) (13, 0.26499998569488525) (14, 0.48399999737739563) (15, 0.41600000858306885) (16, 0.6470000147819519) (17, 0.4059999883174896) (18, 0.9850000143051147) (19, 0.9470000267028809) (20, 0.8939999938011169) (21, 0.8090000152587891) (22, 0.9079999923706055) (23, 0.902999997138977) (24, 0.8389999866485596) (25, 0.8320000171661377) (26, 0.9409999847412109) (27, 0.8949999809265137) (28, 0.9879999756813049) (29, 0.9639999866485596)};
		
		\addplot [	
		draw=brown1,mark=triangle,mark options={brown1,scale=0.3},line width=1pt
		] coordinates {(2, 0.9990000128746033) (3, 0.9890000224113464) (4, 0.9710000157356262) (5, 0.007000000216066837) (6, 0.7210000157356262) (7, 0.9480000138282776) (8, 0.9980000257492065) (9, 0.9440000057220459) (10, 0.9710000157356262) (11, 0.9290000200271606) (12, 0.8700000047683716) (13, 0.8579999804496765) (14, 0.9789999723434448) (15, 0.9670000076293945) (16, 1.0) (17, 0.9819999933242798) (18, 0.9990000128746033) (19, 0.9990000128746033) (20, 0.9959999918937683) (21, 0.9639999866485596) (22, 0.9620000123977661) (23, 0.9879999756813049) (24, 0.9959999918937683) (25, 0.9729999899864197) (26, 0.9980000257492065) (27, 0.9919999837875366) (28, 0.996999979019165) (29, 0.9900000095367432)};
		
		\end{axis}
		
		\end{tikzpicture}%
	\end{subfigure}%
	%\vspace*{-2mm}
	\caption{Accuracy for the selected classes under concept drift. C0 drifts in batches 4 and 5, and C2 drifts in 9 and 10.}
	\label{fig:class_acc}
\end{figure}
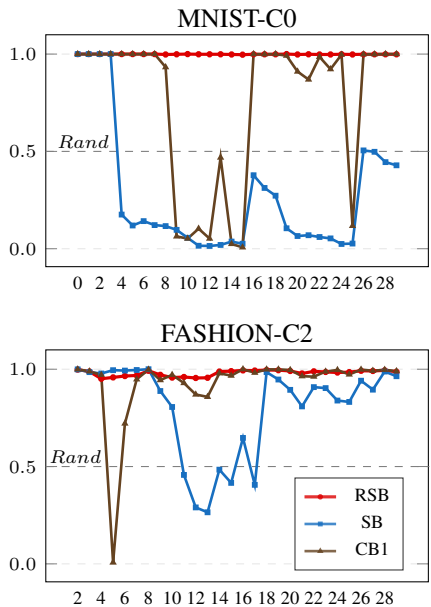

To gain further insights into the performance of the experience replay under concept drift let us look at Fig.~\ref{fig:class_acc} that depicts the accuracy over selected drifting classes. We can see that both CB1 and SB were highly sensitive to any drift in data. Even if sometimes they could spontaneously recover their performance (which usually was rather a coincidence), the next occurrence of concept drift could easily bring their performance back to the level of random decision (or even below). In the case of the MNIST class 0 we can see that the SB method could not recover at any point of time after the first drift. The extremely low accuracy was caused by obsolete centroids, which did not update their label and kept generating invalid instances for the recommendation class. These results clearly indicate that standard experience replay approaches cannot handle concept drifts, and that some of the occurring errors may even never be corrected. On the contrary, the proposed RSB is characterized by excellent robustness to concept drift, stable performance, and on-the-fly adaptation to changes in previously learned classes without any delay or loss in predictive power.  

Finally, we should be aware that concept drift may affect not only the performance of models on previously seen classes, but also their incremental learning capabilities. As the underlying neural network model tries to handle the catastrophic forgetting by using instances from the buffer for experience replay, it utilizes instances coming from outdated concepts that may be contradicting the most current ones. Therefore, this may impact its ability to incorporate and retain new knowledge, resulting in a significant decrease in the model’s predictive power. This allows us to conclude that continual learning under concept drift requires a strong interplay between avoiding catastrophic forgetting and adaptation to concept drift, as weaker performance on one will negatively affect the other. The proposed RSB offers an excellent balance between these two tasks, leading to a well-rounded and stable continual learning solution. 

\section{Conclusions and future works}
\label{sec:con}
\noindent \textbf{Conclusions.} In this paper, we have discussed a unified approach to continual learning that bridges the gap between avoiding catastrophic forgetting and data stream mining under concept drift. By pointing to the fact that these fields are two faces of the same coin, we showed that there is a need for developing holistic systems that are capable of incremental incorporation of new information, while offering adaptation capabilities by selective forgetting. This was illustrated by a practical example of continual learning of user’s preferences that expand and evolve over time. 

To address this challenging scenario, we have proposed an experience replay approach based on a reactive subspace buffer. It combines clustering-driven memory, storing diverse instances per class, with adaptation components that allow for dynamic monitoring, relabeling, and splitting of existing clusters. As a result, our method provides both the capability of accommodating new classes without catastrophic forgetting and the ability to react to concept drift affecting the previously learned classes. In our experimental study, we exhibited the effectiveness of our algorithm and proved that it is an effective and complete approach to continual learning that is not limited by either inability to accommodate new information, or by inability to adapt to changes.

\smallskip
\noindent \textbf{Future works.} We have shown that while existing standard experience replay approaches are able to handle the problem of avoiding catastrophic forgetting, they do not possess mechanisms allowing for adaptation on previously learned classes affected by concept drift. We suppose that similar issues can be identified in other continual learning algorithms. Therefore, our future works will focus on improving different approaches. This may involve, for example, introducing adaptive masking, reactive regularization and dynamic neural network structures capable of reacting to drifts. These will be important steps towards creating a holistic view of continual learning systems that can handle diverse challenges present in various real-life problems. 

{\small
\bibliographystyle{ieee_fullname}
\bibliography{refs}

\begin{thebibliography}{10}\itemsep=-1pt

\bibitem{Aljundi:2019}
Rahaf Aljundi, Klaas Kelchtermans, and Tinne Tuytelaars.
\newblock {Task-Free Continual Learning}.
\newblock In {\em {IEEE} Conference on Computer Vision and Pattern Recognition,
  {CVPR} 2019, Long Beach, CA, USA, June 16-20, 2019}, pages 11254--11263.
  Computer Vision Foundation / {IEEE}, 2019.

\bibitem{Caccia:2020}
Massimo Caccia, Pau Rodriguez, Oleksiy Ostapenko, Fabrice Normandin, Min Lin,
  Lucas Page-Caccia, Issam~Hadj Laradji, Irina Rish, Alexandre Lacoste, David
  V\'{a}zquez, and Laurent Charlin.
\newblock {Online Fast Adaptation and Knowledge Accumulation (OSAKA): a New
  Approach to Continual Learning}.
\newblock In {\em Advances in Neural Information Processing Systems},
  volume~33, pages 16532--16545. Curran Associates, Inc., 2020.

\bibitem{Cano:2020}
Alberto Cano and Bartosz Krawczyk.
\newblock {Kappa Updated Ensemble for drifting data stream mining}.
\newblock {\em Mach. Learn.}, 109(1):175--218, 2020.

\bibitem{Chen:2020}
Hung{-}Jen Chen, An{-}Chieh Cheng, Da{-}Cheng Juan, Wei Wei, and Min Sun.
\newblock {Mitigating Forgetting in Online Continual Learning via
  Instance-Aware Parameterization}.
\newblock In Hugo Larochelle, Marc'Aurelio Ranzato, Raia Hadsell,
  Maria{-}Florina Balcan, and Hsuan{-}Tien Lin, editors, {\em Advances in
  Neural Information Processing Systems 33: Annual Conference on Neural
  Information Processing Systems 2020, NeurIPS 2020, December 6-12, 2020,
  virtual}, 2020.

\bibitem{Barros:2018}
Roberto Souto~Maior de Barros and Silas Garrido~Teixeira de Carvalho~Santos.
\newblock {A large-scale comparison of concept drift detectors}.
\newblock {\em Inf. Sci.}, 451-452:348--370, 2018.

\bibitem{Foerster:2017}
Jakob~N. Foerster, Nantas Nardelli, Gregory Farquhar, Triantafyllos Afouras,
  Philip H.~S. Torr, Pushmeet Kohli, and Shimon Whiteson.
\newblock {Stabilising Experience Replay for Deep Multi-Agent Reinforcement
  Learning}.
\newblock In {\em Proceedings of the 34th International Conference on Machine
  Learning, {ICML} 2017, Sydney, NSW, Australia, 6-11 August 2017}, volume~70
  of {\em Proceedings of Machine Learning Research}, pages 1146--1155. {PMLR},
  2017.

\bibitem{Fujimoto:2020}
Scott Fujimoto, David Meger, and Doina Precup.
\newblock {An Equivalence between Loss Functions and Non-Uniform Sampling in
  Experience Replay}.
\newblock In {\em Advances in Neural Information Processing Systems 33: Annual
  Conference on Neural Information Processing Systems 2020, NeurIPS 2020,
  December 6-12, 2020, virtual}, 2020.

\bibitem{Gou:2020}
Xiangyang Gou, Long He, Yinda Zhang, Ke Wang, Xilai Liu, Tong Yang, Yi Wang,
  and Bin Cui.
\newblock {Sliding Sketches: {A} Framework using Time Zones for Data Stream
  Processing in Sliding Windows}.
\newblock In {\em {KDD} '20: The 26th {ACM} {SIGKDD} Conference on Knowledge
  Discovery and Data Mining, Virtual Event, CA, USA, August 23-27, 2020}, pages
  1015--1025. {ACM}, 2020.

\bibitem{Hayes:2019}
T.~L. {Hayes}, N.~D. {Cahill}, and C. {Kanan}.
\newblock {Memory Efficient Experience Replay for Streaming Learning}.
\newblock In {\em 2019 International Conference on Robotics and Automation
  (ICRA)}, pages 9769--9776, 2019.

\bibitem{Kemker:2018}
Ronald Kemker, Marc McClure, Angelina Abitino, Tyler Hayes, and Christopher
  Kanan.
\newblock {Measuring Catastrophic Forgetting in Neural Networks}.
\newblock {\em Proceedings of the AAAI Conference on Artificial Intelligence},
  32(1), Apr. 2018.

\bibitem{Kirkpatrick:2017}
James Kirkpatrick, Razvan Pascanu, Neil Rabinowitz, Joel Veness, Guillaume
  Desjardins, Andrei~A. Rusu, Kieran Milan, John Quan, Tiago Ramalho, Agnieszka
  Grabska-Barwinska, Demis Hassabis, Claudia Clopath, Dharshan Kumaran, and
  Raia Hadsell.
\newblock {Overcoming catastrophic forgetting in neural networks}.
\newblock {\em Proceedings of the National Academy of Sciences},
  114(13):3521--3526, 2017.

\bibitem{Korycki:2020}
Ł. {Korycki} and B. {Krawczyk}.
\newblock {Online Oversampling for Sparsely Labeled Imbalanced and
  Non-Stationary Data Streams}.
\newblock In {\em 2020 International Joint Conference on Neural Networks
  (IJCNN)}, pages 1--8, 2020.

\bibitem{Krawczyk:2017}
Bartosz Krawczyk, Leandro~L. Minku, Jo{\~{a}}o Gama, Jerzy Stefanowski, and
  Michal Wozniak.
\newblock {Ensemble learning for data stream analysis: {A} survey}.
\newblock {\em Inf. Fusion}, 37:132--156, 2017.

\bibitem{Lee:2016}
Stefan Lee, Senthil Purushwalkam, Michael Cogswell, Viresh Ranjan, David~J.
  Crandall, and Dhruv Batra.
\newblock {Stochastic Multiple Choice Learning for Training Diverse Deep
  Ensembles}.
\newblock In {\em Advances in Neural Information Processing Systems 29: Annual
  Conference on Neural Information Processing Systems 2016, December 5-10,
  2016, Barcelona, Spain}, pages 2119--2127, 2016.

\bibitem{Li:2019}
Xilai Li, Yingbo Zhou, Tianfu Wu, Richard Socher, and Caiming Xiong.
\newblock {Learn to Grow: {A} Continual Structure Learning Framework for
  Overcoming Catastrophic Forgetting}.
\newblock In Kamalika Chaudhuri and Ruslan Salakhutdinov, editors, {\em
  Proceedings of the 36th International Conference on Machine Learning, {ICML}
  2019, 9-15 June 2019, Long Beach, California, {USA}}, volume~97 of {\em
  Proceedings of Machine Learning Research}, pages 3925--3934. {PMLR}, 2019.

\bibitem{Lu:2019}
Jie Lu, Anjin Liu, Fan Dong, Feng Gu, Jo{\~{a}}o Gama, and Guangquan Zhang.
\newblock {Learning under Concept Drift: {A} Review}.
\newblock {\em {IEEE} Trans. Knowl. Data Eng.}, 31(12):2346--2363, 2019.

\bibitem{Mallya:2018}
Arun Mallya, Dillon Davis, and Svetlana Lazebnik.
\newblock {Piggyback: Adapting a Single Network to Multiple Tasks by Learning
  to Mask Weights}.
\newblock In Vittorio Ferrari, Martial Hebert, Cristian Sminchisescu, and Yair
  Weiss, editors, {\em Computer Vision - {ECCV} 2018 - 15th European
  Conference, Munich, Germany, September 8-14, 2018, Proceedings, Part {IV}},
  volume 11208 of {\em Lecture Notes in Computer Science}, pages 72--88.
  Springer, 2018.

\bibitem{Parisi:2019}
German~Ignacio Parisi, Ronald Kemker, Jose~L. Part, Christopher Kanan, and
  Stefan Wermter.
\newblock {Continual lifelong learning with neural networks: {A} review}.
\newblock {\em Neural Networks}, 113:54--71, 2019.

\bibitem{Rolnick:2019}
David Rolnick, Arun Ahuja, Jonathan Schwarz, Timothy~P. Lillicrap, and Gregory
  Wayne.
\newblock {Experience Replay for Continual Learning}.
\newblock In {\em Advances in Neural Information Processing Systems 32: Annual
  Conference on Neural Information Processing Systems 2019, NeurIPS 2019,
  December 8-14, 2019, Vancouver, BC, Canada}, pages 348--358, 2019.

\bibitem{Rostami:2019}
Mohammad Rostami, Soheil Kolouri, and Praveen~K. Pilly.
\newblock {Complementary Learning for Overcoming Catastrophic Forgetting Using
  Experience Replay}.
\newblock In {\em Proceedings of the Twenty-Eighth International Joint
  Conference on Artificial Intelligence, {IJCAI} 2019, Macao, China, August
  10-16, 2019}, pages 3339--3345. ijcai.org, 2019.

\bibitem{Sahoo:2018}
Doyen Sahoo, Quang Pham, Jing Lu, and Steven C.~H. Hoi.
\newblock {Online Deep Learning: Learning Deep Neural Networks on the Fly}.
\newblock In J{\'{e}}r{\^{o}}me Lang, editor, {\em Proceedings of the
  Twenty-Seventh International Joint Conference on Artificial Intelligence,
  {IJCAI} 2018, July 13-19, 2018, Stockholm, Sweden}, pages 2660--2666.
  ijcai.org, 2018.

\bibitem{Schaul:2016}
Tom Schaul, John Quan, Ioannis Antonoglou, and David Silver.
\newblock {Prioritized Experience Replay}.
\newblock In Yoshua Bengio and Yann LeCun, editors, {\em 4th International
  Conference on Learning Representations, {ICLR} 2016, San Juan, Puerto Rico,
  May 2-4, 2016, Conference Track Proceedings}, 2016.

\bibitem{Shi:2018}
Haobin Shi, Shike Yang, Kao{-}Shing Hwang, Jialin Chen, Mengkai Hu, and
  Heng{-}sheng Zhang.
\newblock {A Sample Aggregation Approach to Experiences Replay of Dyna-Q
  Learning}.
\newblock {\em {IEEE} Access}, 6:37173--37184, 2018.

\bibitem{Sokar:2020}
Ghada Sokar, Decebal~Constantin Mocanu, and Mykola Pechenizkiy.
\newblock {SpaceNet: Make Free Space For Continual Learning}.
\newblock {\em CoRR}, abs/2007.07617, 2020.

\bibitem{Sublime:2017}
J{\'{e}}r{\'{e}}mie Sublime, Basarab Matei, and Pierre{-}Alexandre Murena.
\newblock {Analysis of the influence of diversity in collaborative and
  multi-view clustering}.
\newblock In {\em 2017 International Joint Conference on Neural Networks,
  {IJCNN} 2017, Anchorage, AK, USA, May 14-19, 2017}, pages 4126--4133. {IEEE},
  2017.

\bibitem{Oswald:2020}
Johannes von Oswald, Christian Henning, Jo{\~{a}}o Sacramento, and Benjamin~F.
  Grewe.
\newblock {Continual learning with hypernetworks}.
\newblock In {\em 8th International Conference on Learning Representations,
  {ICLR} 2020, Addis Ababa, Ethiopia, April 26-30, 2020}. OpenReview.net, 2020.

\bibitem{Wu:2019}
Yue Wu, Yinpeng Chen, Lijuan Wang, Yuancheng Ye, Zicheng Liu, Yandong Guo, and
  Yun Fu.
\newblock {Large Scale Incremental Learning}.
\newblock In {\em Proceedings of the IEEE/CVF Conference on Computer Vision and
  Pattern Recognition (CVPR)}, June 2019.

\bibitem{Yu:2020}
Lu Yu, Bartlomiej Twardowski, Xialei Liu, Luis Herranz, Kai Wang, Yongmei
  Cheng, Shangling Jui, and Joost van~de Weijer.
\newblock {Semantic Drift Compensation for Class-Incremental Learning}.
\newblock In {\em 2020 {IEEE/CVF} Conference on Computer Vision and Pattern
  Recognition, {CVPR} 2020, Seattle, WA, USA, June 13-19, 2020}, pages
  6980--6989. {IEEE}, 2020.

\bibitem{Zhang:2019}
Mengmi Zhang, Tao Wang, Joo~Hwee Lim, and Jiashi Feng.
\newblock {Prototype Reminding for Continual Learning}.
\newblock {\em CoRR}, abs/1905.09447, 2019.

\bibitem{Zhang:2019sd}
Shiyue Zhang and Mohit Bansal.
\newblock {Addressing Semantic Drift in Question Generation for Semi-Supervised
  Question Answering}.
\newblock In {\em Proceedings of the 2019 Conference on Empirical Methods in
  Natural Language Processing and the 9th International Joint Conference on
  Natural Language Processing, {EMNLP-IJCNLP} 2019, Hong Kong, China, November
  3-7, 2019}, pages 2495--2509. Association for Computational Linguistics,
  2019.

\end{thebibliography}
}

\end{document}